\documentclass[11pt]{article}

\usepackage[preprint]{acl}

\usepackage{times}
\usepackage{latexsym}

\usepackage[T1]{fontenc}

\usepackage[utf8]{inputenc}

\usepackage{microtype}

\usepackage{inconsolata}

\usepackage{graphicx}
\usepackage{enumitem}
\usepackage{booktabs}
\usepackage{makecell}
\usepackage{amsmath}
\usepackage{forest}

\usepackage{tcolorbox}
\tcbuselibrary{listings,breakable}

\newtcblisting{promptbox}[1]{
  listing only, breakable, title={#1},
  fonttitle=\small\bfseries, colback=gray!5, colframe=gray!60,
  listing options={basicstyle=\ttfamily\scriptsize,
                   breaklines=true, breakatwhitespace=true,
                   columns=fullflexible, keepspaces=true,
                   literate={ó}{{\'o}}1 {ç}{{\c{c}}}1}
}

\title{LLMs Are Already Confidence-Aware Zero-Shot Classifiers}
\author{Owen Cook and Xingyi Song \\
University of Sheffield\\
Department of Computer Science\\
\texttt{oscook1@sheffield.ac.uk}, \texttt{x.song@sheffield.ac.uk}}

\begin{document}
\maketitle

\begin{abstract}
Zero-shot large language models (LLMs) are increasingly used as text classifiers, but their labels are only useful when we know to what extent we can trust them. We compare black-box methods (verbalised confidence and sampling) with white-box methods that read token probabilities (P(True), and multiple logit-scoring methods using constrained generation). We evaluate eleven open-weight LLMs (1B--32B parameters) on five classification datasets with 3 to 151 labels, measuring classification performance (accuracy, F1-macro), discrimination (AUROC), and calibration (ECE and Brier score). The choice of confidence method rarely impacts accuracy significantly, but logit-based scoring methods lead to much better discrimination (AUROC 0.74--0.77) than black-box methods (AUROC 0.57--0.66). Replacing labels with single-token identifiers preserves discrimination but costs 3--6 accuracy points on average. We also introduce decision-only scoring, which only scores positions where labels diverge in the prefix tree (trie). It reduces decode steps by 88--96\% on the three largest label sets and performs comparably to full-label scoring. The best open-weight model-method configurations (Gemma~4 31B and Qwen~3.5 9B) are competitive with the commercial Jev 1.13.0 classifier.
\end{abstract}

\section{Introduction}
Large language models (LLMs), now most commonly used as chatbots or coding agents, have a range of other emergent abilities, including classification \citep{wei2022emergent}. As their performance as zero-shot classifiers across a range of domains has improved, LLMs have been increasingly considered for adoption across a number of high-stakes industries such as politics \citep{chatrath2024fact, burnham2026political}, healthcare \citep{artsi2025large, sakai2026large}, and business \citep{li2023chatgpt}. With this, it is increasingly important to understand when we can trust their outputs, and to what extent, using the model's confidence as an indicator.

While there is evidence that smaller fine-tuned language models with a classification head often outperform higher-parameter zero-shot LLMs \citep{bucher2024fine}, fine-tuning such a model is not always possible. Although zero-shot classification with LLMs is not necessarily an optimal solution, it provides more flexibility and can be much more cost-effective, and the gap between zero-shot and fine-tuned classifiers is decreasing. Zero-shot classification can be particularly beneficial in data-scarce environments \citep{zhang2026zero}, where data annotation costs are high, and when label sets may change often, meaning new classifiers do not need to be trained after each change \citep{meyuhas2024iot}. Semi-automated data labelling with LLMs is also growing in popularity \citep{calderon2025alternative, gilardi2023chatgpt}; ensuring the quality of annotations, and good confidence calibration is particularly important in this setting, enabling efficient use of human annotators, without negatively impacting dataset quality \citep{wang2024human}.

Confidence can be extracted from LLMs in multiple ways. Common strategies include asking the model to verbalise its confidence \citep{tian2023just}, sampling multiple responses with a given temperature \citep{wang2022self, xiong2024can}, asking whether its answer is correct \citep{kadavath2022language}, and scoring labels using token probabilities \citep{zhao2021calibrate, fadeeva-etal-2023-lm, kuhn2023semantic}. These confidence extraction methods are rarely compared under the same conditions. The majority of studies investigating confidence extraction focus on question answering (QA) tasks, where knowledge is the key cause of uncertainty. In classification tasks, this is not the case; here uncertainty is much more likely to arise from task ambiguity rather than knowledge issues. In classification problems, labels are often many tokens long and they share decoding paths, so aggregations must represent the full label.
To aggregate label probabilities, outputs must be constrained but the model's first token may be a preamble to the label rather than the answer itself \citep{wang2024my}.

This work provides empirical evidence for the most effective ways of using decoder-only LLMs as zero-shot classifiers, placing particular importance on each method's discriminative abilities and confidence calibration as well as pure classification performance (accuracy, F1-macro). We test using popular classification datasets with labels containing overlapping decoding paths to better test the full-label aggregation methods and ensure they are robust to edge cases; an example of label overlap in SST-5 with the Gemma 4 tokeniser would be: ``very'' ``positive'' and ``very'' ``negative''. For this, we investigate a number of logit-score aggregation methods. 

With the relatively recent release and popularity of Jev-style models, transformer-based classification models that provide a distribution over the whole label set, we use TypeSafe AI's Jev~(1.13.0)\footnote{https://docs.typesafe.ai/models} as a reference point. Zero-shot model-method combinations in this study are competitive with Jev~1.13.0, performing comparably in terms of accuracy and AUROC, but generally worse in terms of raw calibration.

\paragraph{Contributions:}
\begin{itemize}[nosep]
    \item We provide a systematic overview of zero-shot classification with LLMs, focusing on the discrimination and calibration across a range of models for classification tasks.
    \item We introduce a simple scoring rule that only scores and length normalises over decision points in the decode trie, reducing the number of decode steps by 96\% on Banking77, 92\% on CLINC150, and 88\% on 20NG (with the Gemma 4 tokeniser); these reductions in decode steps will not match exactly with the time saved due to overheads.
    \item Experimental code is publicly released on GitHub: \url{https://github.com/minieggz/llm-as-classifier}.
\end{itemize}

\paragraph{Key Findings:}
\begin{itemize}[nosep]
    \item Zero-shot LLMs are natural classifiers that can discriminate relatively well, achieving a mean AUROC of up to 0.802 (Qwen 3.5 9B, constrained with Google length normalisation).
    \item Reducing labels to single tokens often causes lower classification performance but retains discriminative abilities.
    \item Verbalised confidence, or constrained decoding and the choice of scoring methods rarely statistically significantly impact the accuracy of zero-shot LLM classifiers.
    \item Verbalised confidence (on a Likert-scale) achieves the highest calibration overall but is significantly worse in terms of discrimination; methods should not be chosen on calibration alone.
\end{itemize}

\section{Related Work}

\subsection{Constrained Decoding}
As modern LLMs are generally successful at following output constraints, it is common to obtain labels through regex or simple string matching \citep{biderman2024lessons}. Frameworks such as Outlines \citep{willard2023efficient}, Guidance \citep{guidance2022}, and XGrammar \citep{dong2025xgrammar} restrict output by programmatically conforming to a regular expression or grammar, masking disallowed tokens at each decoding step. 
While constrained decoding can ensure we get an answer that fits a specific format, it can distort the LLM's output distribution \citep{park2024grammar}.

\subsection{Extracting Confidence from LLMs}
Confidence can be extracted from LLMs with black-box methods, which have no access to model internals, or white-box methods which have total access.

\paragraph{Unconstrained approaches.}
Instruction tuning has been reported to degrade token probability calibration relative to base variants \citep{kadavath2022language}, motivating the use of verbal approaches \citep{tian2023just}. Confidence can be verbalised either on a Likert-scale or numerically \citep{tian2023just, farr2024llm}. Using an LLM-as-a-judge approach, where the model verbalises its confidence separately to the classification label is also reported to be effective \citep{liu2025uncertainty}.

Another way to calculate confidence is to run inference with a set temperature a number of times; the more times inference is run, the closer the output distribution should be to the model's true answer distribution \citep{wang2022self, xiong2024can}. The majority decision is chosen as the label, with the confidence equal to the proportion of decisions that were the majority label.

The P(True) method \citep{kadavath2022language} uses the answer generated as input to asking the model whether it believes that answer is true or false. The log likelihoods of the first tokens are then softmaxed to obtain the probabilities. 

\paragraph{Scoring multi-token labels.}
Many white-box methods focus on token-level probabilities \citep{geng2024survey}. Confidence is often simply represented by the likelihood of the whole sequence (calculated by summing the log probabilities of each token) \citep{fadeeva-etal-2023-lm, kuhn2023semantic} or a length-normalised likelihood \citep{zhao2021calibrate, gu2025olmes}. These methods are mostly applied to QA datasets. This length normalisation is often simple token length. \citet{wu2016google} apply an alternative length normalisation penalty that has not yet been applied to multi-token label scoring for classification.

For classification tasks specifically, some approaches also normalise the likelihood based on a prior for each label, found by inspecting token probabilities after a blank input \citep{zhao2021calibrate, sanz2025mitigating}. \citet{kumar2024gen} flip this approach with Gen-Z by asking what the likelihood of the input is given each possible output label, removing the requirement of length normalisation due to a constant output size.

A common way to avoid multi-token scoring is to replace full labels with single-token identifiers (often letters), but LLMs can show systematic preferences for particular option identifiers \citep{zheng2024large, pezeshkpour2024large}. Despite this, some open-source Jev-style methods \citep{geuter2026releasing, li2026llmasjevllmsjevstyledecision} take this kind of approach.

TypeSafe AI's Jev API model, trained using Reinforcement Learning for Calibrated Decisions (RLCD) \citep{typesafe2026jev}, supports 255 output classes. It returns a top class label along with a full probability distribution across all labels.

\section{Methodology}
In this study, we consider classification tasks with a static label set. Although each method can be used in cases where the label set is constantly changing, using a static label set simplifies the evaluation and enables us to analyse the performance differences at the class level (F1-macro).

Much of the prior work on short-sequence model confidence calibration focuses on the question answering (QA) task \citep{tian2023just, xiong2024can}. QA relies heavily on recalling knowledge; the classification datasets we use do not rely on recall, so the model's uncertainty reflects the difficulty of assigning a label to a given input, rather than how well the model ``knows'' a fact.

\subsection{Datasets}
We evaluate on text-classification datasets spanning a number of tasks, including topic classification, intent classification, sentiment detection, and fact-checking. We mainly focus on datasets that have labels of varying lengths and that contain overlapping token paths to investigate the issues of length normalisation and the difficulty of extracting full distributions across the label space. These datasets are: Banking77 \citep{casanueva2020efficient}, CLINC150 \citep{larson2019clinc150}, 20 Newsgroups \citep{lang1995newsweeder}, and SST-5 \citep{socher2013recursive}. FEVER \citep{thorne2018fever}, a fact-checking dataset, is used to evaluate cases where all labels diverge at the first token with labels ``SUPPORTS'', ``REFUTES'', and ``NOT ENOUGH INFO''.

Banking77 and CLINC150 are both intent datasets and have the largest number of classes at 77 and 151 (including out-of-scope) respectively, with a number of overlapping tokens within the labels. 20 Newsgroups, a topic classification dataset, contains overlapping labels that are organised into topics and subtopics; labels with the same parent topic will therefore share decoding paths. SST-5 provides a case where the two labels that share token paths are total opposites (``very positive'' and ``very negative'').

\subsection{Large Language Models}
\label{sec:models}
In this study, we use only open-weight language models available on HuggingFace, using the Transformers library \citep{wolf-etal-2020-transformers}. We cover a range of model families and sizes (1B to 32B parameters). These models include Gemma 4 E2B, Gemma 4 E4B, and Gemma 4 31B \citep{team2026gemma}; Llama 3.2 1B and 3B \citep{grattafiori2024llama}; Qwen 2.5 7B \citep{qwen2025qwen25technicalreport}; Qwen 3 8B, Qwen 3 14B, and Qwen 3 32B \citep{yang2025qwen3}; Qwen 3.5 4B, and Qwen 3.5 9B \citep{qwen2026qwen35}. For all models, we are only testing their abilities without chain-of-thought; in most practical classification scenarios, chain-of-thought would significantly increase computational cost.

\subsection{Decoding Methods}
\label{sec:decoding-methods}

Unconstrained decoding represents a simpler chatbot setup where answers are extracted after generation using only the output text. With constrained methods, we access the individual token probabilities at each decode step.

\subsubsection{Unconstrained Generation}
\paragraph{Simple inference methods.}
For the unconstrained generation, we test a number of prompting strategies to extract the answer from the LLM in different formats. These formats include: 1) plaintext label only (``Unconstrained (no conf)''), 2) JSON format with only the label (``Unconstrained JSON (no conf)''), 3) JSON format with confidence before the label (``Unconstrained JSON verbal (pre)''), and 4) JSON format with confidence after the label (``Unconstrained JSON verbal (post)''). We also test single-token substitutions for labels -- ``ABC'' in Table \ref{tab:rq1_summary} -- and Likert-scaled confidence instead of percentage (``Likert''). This Likert scale is ``Not at all confident'', ``Slightly confident'', ``Moderately confident'', ``Very confident'', and ``Completely confident'', equating to 20\%, 40\%, 60\%, 80\%, and 100\%. 

By using the JSON format for confidence extraction, we are using a structure that is easy to obtain both the answer and confidence from and that language models are often post-trained on for tool calling \citep{grattafiori2024llama, yang2025qwen3}. For the plaintext label only, we use regular expressions to extract the largest label first (to prevent cases where ``positive'' may be matched before ``very positive''). We again use regular expressions to extract the JSON object, greedily extracting everything from the first ``\{'' to the final ``\}''. If there is more than one JSON object, this will be picked up as a generation error once parsed. We use the \texttt{json} module to parse JSON output.

\paragraph{Sampling with temperature.}
As an extension to simple plaintext labels, we obtain a confidence estimate by sampling a number of language model outputs using a given temperature score \citep{wang2022self, xiong2024can}. To test this method, we use a
temperature of 0.5, which \citet{kuhn2023semantic} found to be optimal for
sampling-based uncertainty estimation (from a range of 0.25 to 1.5); we sample 10 times per data point. Any output errors here will be removed completely, so if there are 2 errors, we will divide the number of occurrences of each label by 8 rather than 10. We do note that a resolution of 10\% in confidence may impact the results. In all results, this method is referenced as ``Unconstrained (10 samples)''.

\subsubsection{Constrained Decoding}
With constrained decoding, it is possible to obtain a probability distribution over the full label set. We constrain the decoding process using a trie data structure \citep{de2020autoregressive, li2026llmasjevllmsjevstyledecision}.

\paragraph{Tree Construction.}
We represent the constrained decoding process using the trie data structure, using a depth-first approach in its generation. By using the trie, we can optimise the use of the KV-cache for efficient computation. At each decode step, we softmax over the whole vocabulary and then mask the output to only contain the valid next tokens. 

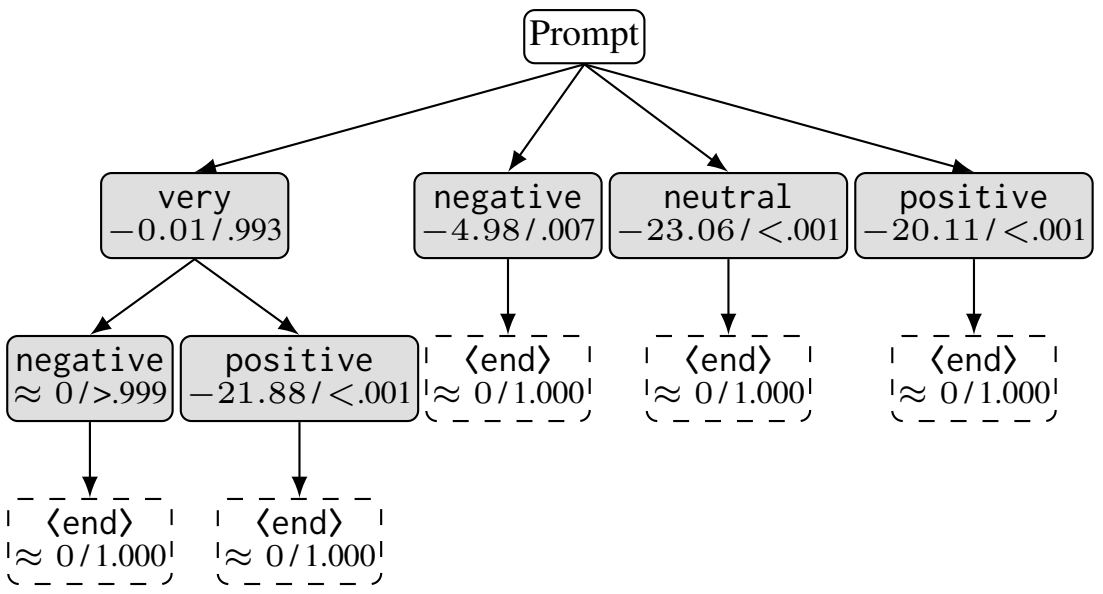
\begin{figure}[t]
\centering
\begin{forest}
for tree={
  font=\scriptsize, align=center, inner sep=1pt,
  rounded corners=2pt, draw, edge={-latex},
  l sep=3.5mm, s sep=0.4mm, parent anchor=south, child anchor=north,
},
root/.style={fill=white},
dec/.style={fill=gray!25},
forced/.style={dashed},
[{Prompt}, root
  [{\texttt{very}\\[-2pt] {\tiny $-0.01$\,/\,.993}}, dec
    [{\texttt{negative}\\[-2pt] {\tiny $\approx 0$\,/\,>.999}}, dec
      [{\texttt{\textlangle end\textrangle}\\[-2pt] {\tiny $\approx 0$\,/\,1.000}}, forced]
    ]
    [{\texttt{positive}\\[-2pt] {\tiny $-21.88$\,/\,$<$.001}}, dec
      [{\texttt{\textlangle end\textrangle}\\[-2pt] {\tiny $\approx 0$\,/\,1.000}}, forced]
    ]
  ]
  [{\texttt{negative}\\[-2pt] {\tiny $-4.98$\,/\,.007}}, dec
    [{\texttt{\textlangle end\textrangle}\\[-2pt] {\tiny $\approx 0$\,/\,1.000}}, forced]
  ]
  [{\texttt{neutral}\\[-2pt] {\tiny $-23.06$\,/\,$<$.001}}, dec
    [{\texttt{\textlangle end\textrangle}\\[-2pt] {\tiny $\approx 0$\,/\,1.000}}, forced]
  ]
  [{\texttt{positive}\\[-2pt] {\tiny $-20.11$\,/\,$<$.001}}, dec
    [{\texttt{\textlangle end\textrangle}\\[-2pt] {\tiny $\approx 0$\,/\,1.000}}, forced]
  ]
]
\end{forest}
\caption{Example trie from SST-5 (Gemma 4 31B). Each node shows the token with \texttt{log-probability / probability}, where the log-probability is over the whole vocabulary and probability is between siblings. Shaded tokens are decision tokens; dashed tokens are forced and are skipped by the decision-only and probability methods. For other datasets, non-decision tokens can be in the middle of labels.}
\label{fig:label-trie}
\end{figure}

A trie is constructed for each prompt and each prompt corresponds to an individual data point. At the end of each label, we calculate the joint likelihood of ``end-of-answer'' tokens. This may be only the EOS token for some models; in other instruction-tuned models, there are specialised end-of-turn tokens (such as \texttt{$<turn|>$} for Gemma 4). For the JSON output specifically, the tokens \texttt{"} and \texttt{"\}} are also added. In the trie, we represent the combined ``end-of-answer'' tokens as token ID ``$-1$''. An example of the trie structure can be seen in Figure \ref{fig:label-trie}. In our experiments, we run trie construction for each model-dataset combination and label aggregation is performed on the saved tries afterwards.

For JSON responses, we prefill the beginning of the JSON output (for example ``\texttt{\{"label": "}''). The exact prefill token sequence was obtained through three example outputs by each model.

\paragraph{Tree aggregation.}
For a label $y$ with tokens $t_1,\dots,t_n$, let
$\ell_i = \log p_\theta(t_i \mid x, t_{<i})$, given input $x$. Most methods assign a score $s(y)$, from which we obtain a distribution over all labels and confidence:
\begin{equation}
\begin{aligned}
\hat{p}(y \mid x) &= \frac{\exp s(y)}{\sum_{y' \in \mathcal{Y}} \exp s(y')}, \\
\mathrm{conf}(x) &= \max_{y \in \mathcal{Y}} \hat{p}(y \mid x).
\end{aligned}
\label{eq:softmax}
\end{equation}
Some methods redefine $\hat{p}(y \mid x)$ directly.

\paragraph{Greedy constrained decoding.}
To directly compare constrained generation against unconstrained generation, we simply use the greedy approach, assigning 100\% confidence to the greedy-decoded output. In this instance $\hat{p}(y \mid x) = 1$ if the greedy path chooses label $y$, otherwise $\hat{p}(y \mid x) = 0$.

\paragraph{Total log likelihood and length normalisation.}
The simplest scoring method is a total log likelihood sum for each label ($s_{\text{TL}}(y)$) \citep{fadeeva-etal-2023-lm}. A simple extension to this is to normalise the label likelihoods by label length in tokens ($s_{\text{LN}}(y)$) \citep{zhao2021calibrate, gu2025olmes}. As well as simple token-length normalisation, we also test the length normalisation penalty proposed by Google for machine translation ($s_{\text{GLN}}(y)$) \citep{wu2016google}. These scoring rules (Equation \ref{eq:scores}) are used in Equation \ref{eq:softmax} to obtain the distribution and confidence.
\begin{equation}
\begin{aligned}
s_{\text{TL}}(y)  &= \textstyle\sum_{i=1}^{n} \ell_i, \\
s_{\text{LN}}(y)  &= s_{\text{TL}}(y) \,/\, n, \\
s_{\text{GLN}}(y) &= s_{\text{TL}}(y) \,\big/\, \bigl(\tfrac{5+n}{6}\bigr)^{\alpha},
\end{aligned}
\label{eq:scores}
\end{equation}
where $n$ is the label sequence length in tokens and in our experiments, $\alpha = 0.6$ following \citet{vaswani2017attention}.

\paragraph{Decision-only methods.}
\label{sec:decision-only}
Decoding each full label, particularly for datasets with large label sets, can be computationally expensive. We propose performing the same calculations introduced above, but applying them to only decision tokens. A decision token is one with siblings, meaning it was not the only choice available (highlighted in Figure~\ref{fig:label-trie}). Let $D(y)$ be the set of positions of the decision tokens on the path to label $y$. The decision-only scoring functions are
\begin{equation}
\begin{aligned}
s^{D}_{\text{TL}}(y)  &= \textstyle\sum_{i \in D(y)} \ell_i, \\
s^{D}_{\text{LN}}(y)  &= s^{D}_{\text{TL}}(y) \,/\, m, \\
s^{D}_{\text{GLN}}(y) &= s^{D}_{\text{TL}}(y) \,\big/\, \bigl(\tfrac{5+m}{6}\bigr)^{\alpha},
\end{aligned}
\label{eq:decision-scores}
\end{equation}
where $m = |D(y)|$ is the number of decision tokens on the path to $y$, and these scores are used in Equation~\ref{eq:softmax} as before. By removing forced tokens, as well as improving computational efficiency, we may also reduce the bias introduced by length normalisation.

\paragraph{Probability tree.}
We could not find any direct examples of probability trees being used to represent full-label decoding for classification. For the probability tree method, rather than softmaxing after all labels have been decoded, we softmax all sibling nodes at each step and multiply the probabilities of each token at the end:
\begin{equation}
\hat{p}(y \mid x) = \prod_{i=1}^{n}
\frac{p_\theta(t_i \mid x, t_{<i})}
     {\sum_{c \in \mathcal{C}(t_{<i})} p_\theta(c \mid x, t_{<i})},
\label{eq:prob-tree}
\end{equation}
where $\mathcal{C}(t_{<i})$ is the set of valid next tokens (siblings of $t_i$) after the previous tokens $t_{<i}$. Figure~\ref{fig:label-trie} visualises the probability tree. When all labels diverge at the first token, the probability tree method reduces to a probability distribution over single-token identifiers. In constrained methods where we replace labels with single tokens (``Constrained (ABC)'' and ``Constrained JSON (ABC)''), we use the probability tree.

The P(True) method also makes use of the probability tree to get the probability of ``True'' vs. ``False'' on already classified data points.

\subsection{Evaluation}
For every method and model-dataset pair we compute accuracy and F1-macro (classification performance), AUROC (discrimination) \citep{hendrycks2016baseline}, and ECE \citep{naeini2015obtaining, guo2017calibration} and top-label Brier score (calibration) \citep{glenn1950verification, gupta2021top} (Brier score applied to the confidence in the predicted label). We also investigate the number of label generation errors per model.

From these results, we report means over all model-dataset pairs for each method. We use the method with the highest mean rank for each metric as the reference and we compare every other method against it using a two-sided Wilcoxon signed-rank test \citep{wilcoxon1945individual} paired over model-dataset pairs, with Holm correction \citep{holm1979simple} within each metric ($\alpha = 0.05$). 
Individual model results (Appendix~\ref{sec:results-by-model}) are reported without tests, as $n = 5$ cannot reach $p < 0.05$.

We then compare the highest performing model-method combinations against Jev~1.13.0 using the same metrics.

\paragraph{Unconstrained evaluation edge cases.}
When evaluating unconstrained outputs, unparsable errors are treated as incorrect answers. In our F1-macro calculations, we assign invalid predictions to incorrect classes as evenly as possible (the gold label is excluded from this assignment) using a round-robin approach. In each round, we prioritise adding incorrect predictions to the largest class to cause a lower impact on the result. When a method does not provide a confidence score, or confidence is unparsable, we assume 100\% certainty. For constrained outputs, this does not apply as it is impossible to obtain a label outside the label set.

\section{Experimental Setup}

\subsection{Key Comparisons}
In our experiments, we test a number of decoding methods on 5 datasets (Banking77, CLINC150, FEVER, SST-5, 20NG) using 11 models (see Section~\ref{sec:models}). Our main comparisons are constrained vs. unconstrained output, single token output vs. full-label output, JSON format vs. plaintext output, verbal confidence vs. sampling confidence vs. logit-based confidence vs. post-hoc P(True) confidence.

The different decoding methods we compare are described in Section \ref{sec:decoding-methods}.

\subsection{Prompting Methodology}
When prompting the models, we follow the COSTAR-A \citep{ohalete2025costar} prompting style. This format includes ``Context'', ``Objective'', ``Style'', ``Tone'', ``Audience'', ``Response'', and ``Answer''. 

For our prompts, we adapt COSTAR-A as follows: style and tone are merged into a single section; audience is omitted; and response is replaced by a ``Format'' section that describes both the structure of the input and the permitted output labels. We also add an ``Input'' section containing the instance to be classified. The ``Context'', ``Style \& Tone'', and ``Format'' sections are included in the system prompt, and the ``Objective'', ``Input'', and ``Answer'' sections form the user prompt which varies per data point.

With modern LLMs proficient in markdown \citep{chen2025mdeval, freeburg2026last} and XML \citep{anthropic2024xml}, all prompts use the markdown format for the above sections and inputs are provided inside XML tags.

Example prompts can be found in Appendix~\ref{sec:prompts}.

\subsection{Model Settings}
For all models, we use 16-bit precision and we use greedy decoding in all instances (other than for the sampling inference method). We use greedy decoding to provide near-deterministic results that may only differ due to hardware-based floating point calculation differences; we generate with a batch size of 1 to mitigate this further and eliminate padding differences.

\section{Results \& Discussion}

\begin{table*}[t]
\centering
\caption{Mean score over all model--dataset pairs (excluding Llama 3.2 1B, Llama 3.2 3B). \textbf{Bold} indicates the best mean. $^{*}$ indicates a method is not significantly different from the method with the best mean rank (two-sided Wilcoxon signed-rank test paired over model-dataset pairs, Holm-adjusted $p > 0.05$).}
\label{tab:rq1_summary}
\resizebox{\textwidth}{!}{%
\begin{tabular}{lrrrrr}
\toprule
Method & Accuracy & F1-macro & AUROC & ECE & Brier \\
\midrule
Unconstrained (no conf) & \textbf{0.724}$^{*}$ & 0.715$^{*}$ & 0.500 & 0.276 & 0.276 \\
Unconstrained (no conf ABC) & 0.671 & 0.663 & 0.500 & 0.329 & 0.329 \\
Unconstrained (10 samples) & \textbf{0.724}$^{*}$ & 0.715$^{*}$ & 0.569 & 0.254 & 0.255 \\
Unconstrained + P(True) & \textbf{0.724}$^{*}$ & 0.715$^{*}$ & 0.741$^{*}$ & 0.227 & 0.235 \\
\midrule
Unconstrained JSON (no conf) & 0.723$^{*}$ & 0.718$^{*}$ & 0.500 & 0.277 & 0.277 \\
Unconstrained JSON (no conf ABC) & 0.686 & 0.680 & 0.500 & 0.314 & 0.314 \\
Unconstrained JSON verbal (pre) & 0.719$^{*}$ & 0.718$^{*}$ & 0.618 & 0.217 & 0.241$^{*}$ \\
Unconstrained JSON verbal (post) & 0.722$^{*}$ & 0.719$^{*}$ & 0.658 & 0.212 & 0.231$^{*}$ \\
Unconstrained JSON verbal (pre, Likert) & 0.721$^{*}$ & 0.716$^{*}$ & 0.606 & 0.167$^{*}$ & 0.222$^{*}$ \\
Unconstrained JSON verbal (post, Likert) & 0.723$^{*}$ & \textbf{0.720}$^{*}$ & 0.636 & \textbf{0.150}$^{*}$ & \textbf{0.202}$^{*}$ \\
Unconstrained JSON verbal (pre, ABC) & 0.679 & 0.676 & 0.592 & 0.262 & 0.280 \\
Unconstrained JSON verbal (post, ABC) & 0.681 & 0.675 & 0.624 & 0.248 & 0.263 \\
\midrule
Constrained (greedy) & 0.723 & 0.714$^{*}$ & 0.500 & 0.277 & 0.277 \\
Constrained (total likelihood) & 0.723$^{*}$ & 0.715$^{*}$ & 0.764$^{*}$ & 0.223 & 0.232 \\
Constrained (total likelihood, decision-only) & 0.723$^{*}$ & 0.715$^{*}$ & 0.764$^{*}$ & 0.223 & 0.232 \\
Constrained (length norm) & 0.723$^{*}$ & 0.715$^{*}$ & 0.759$^{*}$ & 0.213$^{*}$ & 0.227$^{*}$ \\
Constrained (length norm, decision-only) & 0.723$^{*}$ & 0.715$^{*}$ & 0.754 & 0.185$^{*}$ & 0.211$^{*}$ \\
Constrained (Google length norm) & 0.723$^{*}$ & 0.715$^{*}$ & 0.764$^{*}$ & 0.208 & 0.222 \\
Constrained (Google length norm, decision-only) & 0.723$^{*}$ & 0.715$^{*}$ & 0.763$^{*}$ & 0.220 & 0.229 \\
Constrained (probability tree) & 0.723$^{*}$ & 0.714$^{*}$ & 0.763$^{*}$ & 0.223 & 0.232 \\
Constrained (ABC) & 0.667 & 0.662 & 0.758$^{*}$ & 0.256 & 0.266 \\
\midrule
Constrained JSON (greedy) & 0.715 & 0.707 & 0.500 & 0.285 & 0.285 \\
Constrained JSON (total likelihood) & 0.713 & 0.706 & 0.760$^{*}$ & 0.222 & 0.234 \\
Constrained JSON (total likelihood, decision-only) & 0.716 & 0.707 & 0.763$^{*}$ & 0.222 & 0.234 \\
Constrained JSON (length norm) & 0.681 & 0.679 & 0.738 & 0.249$^{*}$ & 0.264$^{*}$ \\
Constrained JSON (length norm, decision-only) & 0.715 & 0.708 & 0.746$^{*}$ & 0.185$^{*}$ & 0.213$^{*}$ \\
Constrained JSON (Google length norm) & 0.708 & 0.702 & 0.760$^{*}$ & 0.203 & 0.225 \\
Constrained JSON (Google length norm, decision-only) & 0.716 & 0.708 & 0.762$^{*}$ & 0.218 & 0.231 \\
Constrained JSON (probability tree) & 0.715 & 0.707 & 0.762$^{*}$ & 0.222 & 0.234 \\
Constrained JSON (ABC) & 0.683 & 0.680 & \textbf{0.766}$^{*}$ & 0.247 & 0.254 \\
\bottomrule
\end{tabular}}
\end{table*}

\begin{table*}[t]
\centering
\caption{Best model--method combinations for each metric, best constrained method (accuracy), best ABC-logit method (accuracy), compared against Jev 1.13.0. Scores are means across all datasets.}
\label{tab:comparison}
\begin{tabular}{llrrrrr}
\toprule
Model & Method & Accuracy & F1-macro & AUROC & ECE & Brier \\
\midrule
Gemma 4 31B & \makecell[l]{Unconstrained JSON verbal\\(pre, Likert)} & \textbf{0.803} & \textbf{0.796} & 0.642 & 0.120 & 0.145\\
Gemma 4 31B & \makecell[l]{Unconstrained JSON verbal\\(post, Likert)} & 0.802 & \textbf{0.796} & 0.673 & 0.107 & 0.140\\

Qwen 3.5 9B & \makecell[l]{Constrained\\(Google length norm)} & 0.750 & 0.741 & \textbf{0.802} & 0.107 & 0.153\\
Qwen 3.5 9B & \makecell[l]{Constrained\\(length norm, decision-only)} & 0.751 & 0.744 & 0.793 & \textbf{0.053} & 0.142\\
\midrule
Gemma 4 31B & \makecell[l]{Constrained JSON\\(Google length norm,\\decision-only)} & 0.802 & 0.795 & 0.775 & 0.190 & 0.190\\
\midrule
Gemma 4 31B & \makecell[l]{Constrained JSON\\(ABC)} & 0.798 & 0.790 & 0.772 & 0.196 & 0.196\\
\midrule
Jev 1.13.0 & Jev & 0.788 & 0.780 & 0.801 & 0.086 & \textbf{0.134} \\
\bottomrule
\end{tabular}
\end{table*}

\paragraph{Smaller models generally produce more generation errors.}
\begin{figure}[h]
\includegraphics[width=0.47\textwidth]{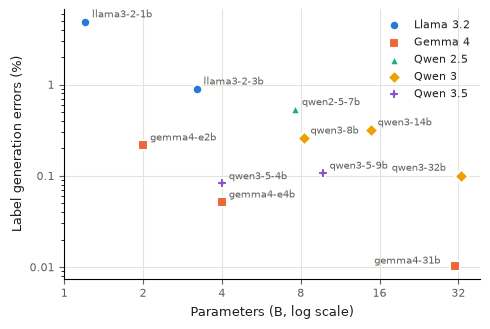}
\caption{Rate of errors made by each model (all unconstrained inference methods: plaintext, JSON with no confidence, verbal pre and post, Likert and percentage) plotted against number of model parameters (in billions) and grouped by model family.}
\label{fig:model_errors}
\end{figure}

As shown in Figure \ref{fig:model_errors}, the number of generation errors generally increases as the number of model parameters decreases. While the general trend is clear from looking at all models, it is most visible within the Gemma 4 family of models where the error rate decreases from 0.22\% with Gemma 4 E2B to 0.01\% for the 31 billion parameter variant. The mid-sized Qwen 3 and Qwen 3.5 models, however, show that this is not always the case.

\paragraph{Despite constrained decoding preventing generation errors, it can harm model performance.}
It is well-investigated that constrained decoding can often distort an LLM's output distribution, in that it is forced to choose between a set of tokens where all of them have extremely low likelihoods \citep{wang2024my}. The alternative formats the unconstrained models used include bullet pointing, prepending with ``Answer: '', and using markdown formatting. For the full results, see Table~\ref{tab:constrained-plaintext} and Table~\ref{tab:constrained-json}.

\paragraph{The confidence extraction method rarely significantly impacts classification performance.}
Post-hoc confidence methods like P(True) do not change the accuracy at all, as they are extensions of greedy inference with no confidence. Other greedy generation methods all use JSON-formatted outputs, and the accuracy is only statistically significantly worse than the best performing method (unconstrained output with no confidence) when labels are reduced to single-token replacements. The averaged constrained JSON methods perform statistically worse than the best performing method, but some of the highest performing model-method combinations do use constrained JSON output, indicating that the preferred format is model-dependent (see Table~\ref{tab:comparison}).

\paragraph{In terms of raw calibration, verbal confidence performs well. Likert-scale confidence is better calibrated but leads to worse discriminative abilities.} For verbal confidence, asking models for JSON output with answer first and confidence second performed the best in terms of ECE and top-label Brier score (shown in Table~\ref{tab:rq1_summary}). While asking for confidence in a Likert scale provides better raw calibration, it leads to a poorer AUROC, most likely due to the significantly lower resolution of the Likert scale compared to percentage output.

\paragraph{Methods that reduce the labels to single tokens often lead to reductions in accuracy and calibration, but perform well discriminatively and are more computationally efficient.}
Overall (see Table \ref{tab:rq1_summary}), all methods that reduce labels to single tokens perform significantly worse than the highest ranking method on average in terms of accuracy, F1-macro, ECE and top-label Brier score. While the accuracy is lower, they perform well on discrimination (AUROC) with ``Constrained JSON (ABC)'' achieving the highest average score. The constrained single-token methods also require only one decode step per data point, making these methods quick and resource-efficient when compared to methods requiring the full decoding of each label.

\paragraph{Tree-based decoding and logit scoring lead to significantly better classifiers in terms of discrimination.}
As shown in Table \ref{tab:rq1_summary}, the constrained logit-based methods perform well, with only two trie-based methods (excluding ``greedy'') performing statistically significantly worse on average than the best performing method ``Constrained JSON (ABC)'' in terms of AUROC. While calibration scores in logit-based methods are not quite as low as some of the verbalised confidence methods overall, the best performing model and method combination for ECE is Qwen 3.5 9B constrained with length normalisation (decision tokens only). Calibration scores could be improved using a small validation set and temperature scaling \citep{guo2017calibration}.

The P(True) method, which is also a logit-based method, is not statistically significantly worse than other logit-based methods in terms of discrimination and maintains the benefits of unconstrained generation for classification performance. Using P(True), however, does not produce a confidence distribution over all labels as other logit-scoring methods do.

\paragraph{Open-weight instruction-tuned models are competitive with Jev 1.13.0.}
Of the best performing methods in Table \ref{tab:comparison}, Jev 1.13.0 only wins on the top-label Brier score but no model-method combination beats Jev on classification performance and discrimination combined. The model-method combination with the highest accuracy and F1-macro is Gemma 4 31B using unconstrained JSON generation with Likert-scale confidence; this does come at a large cost in discrimination. Gemma 4 31B constrained with JSON output and Google length normalisation on decision tokens only, as well as Gemma 4 31B with constrained single-token JSON output, achieves an accuracy and F1-macro higher than Jev, but performs worse in terms of AUROC, ECE, and top-label Brier. The best performing Qwen 3.5 9B method combinations perform similarly or better in terms of AUROC but do lose on raw classification performance. While the Jev API may be a viable option for those without extensive privacy concerns, open-weight alternatives with no extra training are competitive.

\section{Conclusion}
Open-weight zero-shot LLMs are effective classifiers that are competitive with current commercial classification models in terms of accuracy and discrimination. While the raw calibration is not always as strong, it could likely be improved with temperature scaling \citep{guo2017calibration}. Unconstrained outputs using verbalised confidence have been shown to perform well in terms of accuracy, but methods that use logit scores are shown to be significantly better at providing confidence scores that discriminate between correct and incorrect answers. 

Open-weight models can perform very well in terms of classification score, discrimination, and calibration using logit-based methods. Of the models tested, Gemma 4 31B is the most accurate (constrained and unconstrained), but the relatively small Qwen 3.5 9B achieved the best AUROC and ECE scores, with a slightly lower, but still competitive, accuracy and F1-macro of $\approx0.75$ and $\approx0.74$.

The decision for which model and method to use likely depends on hardware, time or throughput, and privacy requirements. If there are no privacy requirements, commercial models like Jev can be a good option for quick, high-quality classification results; there is also no need for a GPU. If time and throughput are not a large concern, decoding full labels can be beneficial; depending on the label set, this can be more costly. By reducing computations to decision-only steps (see Section~\ref{sec:decision-only}), it is possible to save computation, achieving similar results to full decoding. If time is limited and/or there is a very high throughput required, reducing the label set to single tokens can save a lot of time, but it does come at the expense of accuracy. Depending on the model, it may be better to constrain the output to JSON or natural text; a small validation set may be used to test this and which trie-based method to use.

\section*{Limitations}
Jev's architecture and training data are not publicly documented, so we cannot be sure whether or not it was trained on the public datasets we have used. The same applies to the open-weight models.

In this study, we do not investigate the impact of temperature scaling on the calibration of our logit-based methods. Temperature scaling \citep{guo2017calibration}, fit on a small labelled validation set, is likely to improve calibration (ECE and top-label Brier score), while maintaining the same discriminative abilities.

All experiments are zero-shot only. Few-shot prompting may improve classification results further but come at some computational cost. Investigation of this trade-off could be interesting.

Our experiments use a single prompt per method-dataset pair, meaning we do not test for prompt variability. All datasets and prompts are also in English, meaning we cannot confidently draw conclusions on zero-shot LLM classification in other languages, although we expect the same conclusions to hold.

For the ``Google length norm'' methods, we do not tune the $\alpha$ parameter. This could potentially affect classification accuracy, discrimination, and calibration. Sampling also only uses a single setting: 10 samples at temperature 0.5. Testing sampling with a higher resolution and different temperatures could improve calibration for this method.

Experiments use 16-bit precision across four GPU types. Despite greedy decoding and a batch size of 1, floating-point differences could cause small variations in results.

\section*{Acknowledgments}
We acknowledge the IT Services at The University of Sheffield, utilising their High Performance Computing clusters in our experiments.
\bibliography{custom}

\appendix
\label{sec:appendix}

\section{Additional Experimental Details}
\subsection{Dataset Usage}
In all cases other than for Banking77, all dataset labels are left untouched. For Banking77, we changed ``Refund\_not\_showing\_up'' to ``refund\_not\_showing\_up'' and ``reverted\_card\_payment?'' to ``reverted\_card\_payment'' to be more consistent with the rest of the formatting of the labels, as these are known mistakes within the label set. We expect label outputs to exactly match those of the dataset to be considered correct. For all other datasets, we left label names untouched. For example, all FEVER labels remain in upper case and underscores or dots are left between categories and words for other datasets.

To ensure that out-of-memory errors never occur, we only use examples with an input text of 2048 tokens or under. Only FEVER and 20 Newsgroups are affected, with FEVER reduced from 9,999 examples to 9,975 and 20 Newsgroups reduced from 7,317 to 7,121.

\subsection{Hardware}
Experiments were run across 4 different GPU models: NVIDIA A100 with 40GB of VRAM, NVIDIA A100 with 80GB of VRAM, NVIDIA H100 with 80GB of VRAM, and the NVIDIA GH200 with $\sim$96GB HBM3.

\section{Constrained vs. Unconstrained}

\begin{table*}[t]
    \centering
    \small
    \caption{The percentage of data points on which constrained plaintext generation helps ($+$) (constrained output is correct, unconstrained output is either incorrect or unparsable) and hinders ($-$) (unconstrained output is correct, constrained output is incorrect) classification performance by model and dataset. Both constrained and unconstrained generation use greedy decoding.}
    \label{tab:constrained-plaintext}
    \begin{tabular}{lccccc}
        \toprule
         & Banking77 & CLINC150 & FEVER & SST-5 & 20NG \\
        \midrule
        Llama 3.2 1B & +1.66\% / -11.59\% & +3.00\% / -6.07\% & +1.13\% / -0.19\% & +0.00\% / -0.00\% & +0.10\% / -0.13\% \\
        Llama 3.2 3B & +0.55\% / -0.39\% & +2.56\% / -1.82\% & +0.59\% / -0.15\% & +0.09\% / -0.18\% & +0.21\% / -0.13\% \\
        Gemma 4 E2B & +0.16\% / -0.03\% & +0.31\% / -2.78\% & +0.00\% / -0.10\% & +0.00\% / -0.00\% & +0.11\% / -0.83\% \\
        Gemma 4 E4B & +0.03\% / -0.03\% & +0.04\% / -0.02\% & +0.00\% / -0.00\% & +0.00\% / -0.00\% & +0.00\% / -0.00\% \\
        Gemma 4 31B & +0.00\% / -0.00\% & +0.04\% / -0.00\% & +0.01\% / -0.01\% & +0.00\% / -0.05\% & +0.03\% / -0.00\% \\
        Qwen2.5 7B & +0.26\% / -0.36\% & +0.05\% / -0.05\% & +0.15\% / -0.06\% & +0.05\% / -0.00\% & +0.07\% / -0.00\% \\
        Qwen3 8B & +0.03\% / -0.00\% & +0.02\% / -0.05\% & +0.00\% / -0.00\% & +0.00\% / -0.05\% & +0.06\% / -0.01\% \\
        Qwen3 14B & +0.00\% / -0.03\% & +0.15\% / -0.09\% & +0.00\% / -0.00\% & +0.00\% / -0.00\% & +0.10\% / -0.00\% \\
        Qwen3 32B & +0.10\% / -0.00\% & +0.05\% / -0.09\% & +0.00\% / -0.00\% & +0.05\% / -0.00\% & +0.07\% / -0.01\% \\
        Qwen3.5 4B & +0.13\% / -0.00\% & +0.04\% / -0.02\% & +0.00\% / -0.00\% & +0.05\% / -0.00\% & +0.04\% / -0.04\% \\
        Qwen3.5 9B & +0.03\% / -0.00\% & +0.00\% / -0.02\% & +0.01\% / -0.01\% & +0.05\% / -0.00\% & +0.00\% / -0.06\% \\
        \bottomrule
    \end{tabular}
\end{table*}

\begin{table*}[t]
    \centering
    \small
    \caption{The percentage of data points on which constrained JSON generation helps ($+$) (constrained output is correct, unconstrained output is either incorrect or unparsable) and hinders ($-$) (unconstrained output is correct, constrained output is incorrect) classification performance by model and dataset. Both constrained and unconstrained generation use greedy decoding.}
    \label{tab:constrained-json}
    \resizebox{\textwidth}{!}{%
    \begin{tabular}{lccccc}
        \toprule
         & Banking77 & CLINC150 & FEVER & SST-5 & 20NG \\
        \midrule
        Llama 3.2 1B & +4.22\% / -20.13\% & +18.20\% / -8.62\% & +32.25\% / -35.92\% & +11.22\% / -24.75\% & +3.93\% / -3.13\% \\
        Llama 3.2 3B & +2.63\% / -1.40\% & +3.11\% / -3.56\% & +4.45\% / -1.48\% & +2.44\% / -3.57\% & +2.09\% / -5.25\% \\
        Gemma 4 E2B & +0.75\% / -0.91\% & +0.58\% / -0.91\% & +1.01\% / -7.62\% & +3.35\% / -5.29\% & +1.24\% / -2.27\% \\
        Gemma 4 E4B & +0.29\% / -0.49\% & +0.62\% / -3.98\% & +0.33\% / -1.00\% & +1.49\% / -1.86\% & +0.59\% / -1.28\% \\
        Gemma 4 31B & +0.29\% / -0.19\% & +0.09\% / -0.13\% & +0.11\% / -0.06\% & +1.49\% / -0.81\% & +0.10\% / -0.15\% \\
        Qwen2.5 7B & +1.92\% / -2.79\% & +3.56\% / -3.80\% & +0.93\% / -0.86\% & +1.76\% / -2.81\% & +1.80\% / -1.57\% \\
        Qwen3 8B & +2.11\% / -1.17\% & +0.91\% / -3.15\% & +0.33\% / -0.78\% & +4.66\% / -5.02\% & +1.22\% / -2.36\% \\
        Qwen3 14B & +0.81\% / -1.07\% & +0.89\% / -3.27\% & +0.11\% / -1.24\% & +7.47\% / -8.91\% & +1.33\% / -2.37\% \\
        Qwen3 32B & +1.30\% / -1.33\% & +1.95\% / -2.35\% & +0.83\% / -0.76\% & +3.26\% / -2.13\% & +1.57\% / -1.90\% \\
        Qwen3.5 4B & +1.17\% / -1.66\% & +1.55\% / -2.07\% & +0.16\% / -1.04\% & +2.08\% / -3.35\% & +1.67\% / -1.85\% \\
        Qwen3.5 9B & +0.78\% / -0.94\% & +2.16\% / -1.27\% & +0.05\% / -2.28\% & +3.98\% / -6.56\% & +1.24\% / -1.29\% \\
        \bottomrule
    \end{tabular}}
\end{table*}

\clearpage
\section{Results By Model}
\label{sec:results-by-model}

\begin{table*}[t]
\centering
\caption{Accuracy (mean score) across datasets, by model. \textbf{Bold} score represents the best performing for that model. Significance not tested as there are too few cells for the test to reach significance.}
\label{tab:best_model_accuracy}
\resizebox{\textwidth}{!}{%
\begin{tabular}{lrrrrrrrrrrr}
\toprule
Method & Llama 3.2 1B & Llama 3.2 3B & Gemma 4 E2B & Gemma 4 E4B & Gemma 4 31B & Qwen2.5 7B & Qwen3 8B & Qwen3 14B & Qwen3 32B & Qwen3.5 4B & Qwen3.5 9B \\
\midrule
Unconstrained (no conf) & 0.245 & 0.487 & 0.645 & 0.728 & 0.801 & 0.681 & 0.706 & 0.731 & 0.749 & 0.720 & 0.751 \\
Unconstrained (no conf ABC) & 0.162 & 0.341 & 0.552 & 0.677 & 0.795 & 0.607 & 0.619 & 0.702 & 0.716 & 0.658 & 0.710 \\
Unconstrained (10 samples) & 0.245 & 0.490 & 0.646 & \textbf{0.729} & 0.801 & 0.680 & 0.707 & 0.731 & 0.749 & 0.720 & 0.751 \\
Unconstrained + P(True) & 0.245 & 0.487 & 0.645 & 0.728 & 0.801 & 0.681 & 0.706 & 0.731 & 0.749 & 0.720 & 0.751 \\
\midrule
Unconstrained JSON (no conf) & 0.273 & 0.515 & 0.656 & 0.725 & 0.800 & 0.669 & 0.709 & \textbf{0.733} & 0.741 & 0.718 & 0.752 \\
Unconstrained JSON (no conf ABC) & 0.160 & 0.421 & 0.604 & 0.694 & 0.798 & 0.604 & 0.658 & 0.715 & 0.720 & 0.669 & 0.713 \\
Unconstrained JSON verbal (pre) & \textbf{0.295} & 0.506 & 0.657 & 0.716 & 0.800 & 0.665 & 0.707 & 0.711 & 0.741 & \textbf{0.724} & \textbf{0.753} \\
Unconstrained JSON verbal (post) & 0.270 & 0.493 & 0.653 & 0.720 & 0.802 & 0.671 & \textbf{0.716} & 0.729 & 0.743 & 0.720 & 0.747 \\
Unconstrained JSON verbal (pre, Likert) & 0.185 & 0.514 & \textbf{0.661} & 0.724 & \textbf{0.803} & 0.674 & 0.701 & 0.725 & 0.737 & 0.714 & 0.750 \\
Unconstrained JSON verbal (post, Likert) & 0.258 & 0.494 & 0.657 & 0.719 & 0.802 & 0.674 & 0.712 & 0.730 & 0.747 & 0.718 & 0.746 \\
Unconstrained JSON verbal (pre, ABC) & 0.101 & 0.412 & 0.603 & 0.689 & 0.796 & 0.596 & 0.637 & 0.701 & 0.712 & 0.668 & 0.706 \\
Unconstrained JSON verbal (post, ABC) & 0.142 & 0.387 & 0.606 & 0.690 & 0.797 & 0.596 & 0.641 & 0.714 & 0.714 & 0.668 & 0.705 \\
\midrule
Constrained (greedy) & 0.220 & 0.490 & 0.639 & 0.728 & 0.801 & 0.681 & 0.706 & 0.731 & 0.749 & 0.721 & 0.751 \\
Constrained (total likelihood) & 0.220 & 0.494 & 0.639 & \textbf{0.729} & 0.801 & \textbf{0.682} & 0.706 & 0.731 & \textbf{0.750} & 0.722 & 0.751 \\
Constrained (total likelihood, decision-only) & 0.220 & 0.494 & 0.639 & 0.728 & 0.801 & \textbf{0.682} & 0.706 & 0.731 & \textbf{0.750} & 0.721 & 0.751 \\
Constrained (length norm) & 0.182 & 0.484 & 0.636 & 0.727 & 0.801 & \textbf{0.682} & 0.707 & 0.731 & 0.749 & 0.721 & 0.752 \\
Constrained (length norm, decision-only) & 0.218 & 0.490 & 0.636 & 0.727 & 0.801 & 0.681 & 0.706 & 0.731 & 0.749 & 0.721 & 0.751 \\
Constrained (Google length norm) & 0.210 & 0.493 & 0.639 & 0.728 & 0.801 & \textbf{0.682} & 0.706 & 0.731 & \textbf{0.750} & 0.722 & 0.750 \\
Constrained (Google length norm, decision-only) & 0.220 & 0.494 & 0.639 & \textbf{0.729} & 0.801 & \textbf{0.682} & 0.706 & 0.731 & \textbf{0.750} & 0.722 & 0.751 \\
Constrained (probability tree) & 0.220 & 0.493 & 0.639 & 0.728 & 0.801 & 0.681 & 0.706 & 0.731 & \textbf{0.750} & 0.721 & 0.751 \\
Constrained (ABC) & 0.191 & 0.341 & 0.545 & 0.671 & 0.795 & 0.597 & 0.616 & 0.702 & 0.718 & 0.653 & 0.710 \\
\midrule
Constrained JSON (greedy) & 0.228 & 0.514 & 0.636 & 0.715 & 0.802 & 0.665 & 0.703 & 0.721 & 0.742 & 0.711 & 0.744 \\
Constrained JSON (total likelihood) & 0.110 & 0.500 & 0.636 & 0.696 & 0.802 & 0.665 & 0.703 & 0.720 & 0.742 & 0.713 & 0.744 \\
Constrained JSON (total likelihood, decision-only) & 0.230 & 0.517 & 0.636 & 0.715 & 0.802 & 0.666 & 0.703 & 0.722 & 0.742 & 0.712 & 0.744 \\
Constrained JSON (length norm) & 0.115 & 0.462 & 0.636 & 0.508 & 0.802 & 0.664 & 0.640 & 0.681 & 0.743 & 0.712 & 0.744 \\
Constrained JSON (length norm, decision-only) & 0.209 & 0.506 & 0.636 & 0.713 & 0.802 & 0.664 & 0.704 & 0.721 & 0.740 & 0.711 & 0.744 \\
Constrained JSON (Google length norm) & 0.116 & 0.495 & 0.636 & 0.655 & 0.802 & 0.665 & 0.703 & 0.715 & 0.743 & 0.712 & 0.744 \\
Constrained JSON (Google length norm, decision-only) & 0.238 & 0.516 & 0.636 & 0.715 & 0.802 & 0.665 & 0.703 & 0.723 & 0.742 & 0.712 & 0.744 \\
Constrained JSON (probability tree) & 0.227 & \textbf{0.518} & 0.636 & 0.715 & 0.802 & 0.665 & 0.703 & 0.721 & 0.742 & 0.712 & 0.744 \\
Constrained JSON (ABC) & 0.114 & 0.422 & 0.595 & 0.683 & 0.798 & 0.603 & 0.656 & 0.715 & 0.718 & 0.668 & 0.714 \\
\bottomrule
\end{tabular}}
\end{table*}

\begin{table*}[t]
\centering
\caption{F1-macro (mean score) across datasets, by model. \textbf{Bold} score represents the best performing for that model. Significance not tested as there are too few cells for the test to reach significance.}
\label{tab:best_model_f1_macro}
\resizebox{\textwidth}{!}{%
\begin{tabular}{lrrrrrrrrrrr}
\toprule
Method & Llama 3.2 1B & Llama 3.2 3B & Gemma 4 E2B & Gemma 4 E4B & Gemma 4 31B & Qwen2.5 7B & Qwen3 8B & Qwen3 14B & Qwen3 32B & Qwen3.5 4B & Qwen3.5 9B \\
\midrule
Unconstrained (no conf) & 0.175 & 0.470 & 0.637 & \textbf{0.723} & 0.795 & 0.665 & 0.693 & 0.732 & 0.746 & 0.701 & 0.741 \\
Unconstrained (no conf ABC) & 0.134 & 0.315 & 0.546 & 0.681 & 0.789 & 0.588 & 0.603 & 0.695 & 0.710 & 0.648 & 0.703 \\
Unconstrained (10 samples) & 0.189 & 0.478 & 0.637 & \textbf{0.723} & 0.795 & 0.665 & 0.694 & 0.732 & 0.746 & 0.700 & 0.741 \\
Unconstrained + P(True) & 0.175 & 0.470 & 0.637 & \textbf{0.723} & 0.795 & 0.665 & 0.693 & 0.732 & 0.746 & 0.701 & 0.741 \\
\midrule
Unconstrained JSON (no conf) & 0.235 & 0.503 & 0.661 & 0.721 & 0.793 & 0.663 & 0.701 & \textbf{0.738} & 0.739 & 0.697 & \textbf{0.748} \\
Unconstrained JSON (no conf ABC) & 0.098 & 0.389 & 0.609 & 0.698 & 0.790 & 0.590 & 0.650 & 0.710 & 0.715 & 0.657 & 0.703 \\
Unconstrained JSON verbal (pre) & \textbf{0.277} & 0.481 & 0.661 & 0.720 & 0.795 & 0.662 & 0.702 & 0.724 & 0.743 & \textbf{0.711} & \textbf{0.748} \\
Unconstrained JSON verbal (post) & 0.253 & 0.476 & 0.658 & 0.717 & \textbf{0.796} & 0.666 & \textbf{0.706} & 0.735 & 0.742 & 0.706 & 0.743 \\
Unconstrained JSON verbal (pre, Likert) & 0.135 & 0.495 & \textbf{0.664} & 0.722 & \textbf{0.796} & 0.664 & 0.693 & 0.730 & 0.739 & 0.691 & 0.746 \\
Unconstrained JSON verbal (post, Likert) & 0.230 & 0.479 & 0.663 & 0.717 & \textbf{0.796} & \textbf{0.672} & 0.705 & 0.734 & \textbf{0.747} & 0.707 & 0.742 \\
Unconstrained JSON verbal (pre, ABC) & 0.059 & 0.373 & 0.609 & 0.698 & 0.790 & 0.581 & 0.633 & 0.702 & 0.709 & 0.656 & 0.701 \\
Unconstrained JSON verbal (post, ABC) & 0.103 & 0.358 & 0.612 & 0.695 & 0.791 & 0.579 & 0.632 & 0.708 & 0.709 & 0.655 & 0.697 \\
\midrule
Constrained (greedy) & 0.150 & 0.476 & 0.632 & \textbf{0.723} & 0.795 & 0.665 & 0.693 & 0.732 & \textbf{0.747} & 0.701 & 0.741 \\
Constrained (total likelihood) & 0.144 & 0.481 & 0.632 & \textbf{0.723} & 0.795 & 0.667 & 0.694 & 0.732 & \textbf{0.747} & 0.702 & 0.741 \\
Constrained (total likelihood, decision-only) & 0.148 & 0.481 & 0.632 & \textbf{0.723} & 0.795 & 0.666 & 0.694 & 0.732 & \textbf{0.747} & 0.702 & 0.741 \\
Constrained (length norm) & 0.116 & 0.469 & 0.629 & 0.722 & 0.795 & 0.667 & 0.695 & 0.733 & \textbf{0.747} & 0.703 & 0.743 \\
Constrained (length norm, decision-only) & 0.153 & 0.479 & 0.629 & 0.722 & 0.795 & 0.666 & 0.694 & 0.733 & \textbf{0.747} & 0.705 & 0.744 \\
Constrained (Google length norm) & 0.137 & 0.480 & 0.631 & \textbf{0.723} & 0.795 & 0.666 & 0.694 & 0.732 & \textbf{0.747} & 0.703 & 0.741 \\
Constrained (Google length norm, decision-only) & 0.150 & 0.482 & 0.632 & \textbf{0.723} & 0.795 & 0.666 & 0.694 & 0.732 & \textbf{0.747} & 0.703 & 0.742 \\
Constrained (probability tree) & 0.149 & 0.480 & 0.632 & \textbf{0.723} & 0.795 & 0.665 & 0.694 & 0.732 & \textbf{0.747} & 0.702 & 0.741 \\
Constrained (ABC) & 0.151 & 0.316 & 0.543 & 0.679 & 0.789 & 0.587 & 0.601 & 0.695 & 0.713 & 0.646 & 0.703 \\
\midrule
Constrained JSON (greedy) & 0.150 & 0.507 & 0.640 & 0.706 & 0.795 & 0.654 & 0.687 & 0.725 & 0.738 & 0.687 & 0.731 \\
Constrained JSON (total likelihood) & 0.068 & 0.488 & 0.640 & 0.692 & 0.795 & 0.655 & 0.689 & 0.729 & 0.738 & 0.689 & 0.731 \\
Constrained JSON (total likelihood, decision-only) & 0.148 & 0.510 & 0.640 & 0.707 & 0.795 & 0.655 & 0.688 & 0.725 & 0.738 & 0.689 & 0.731 \\
Constrained JSON (length norm) & 0.059 & 0.435 & 0.640 & 0.510 & 0.795 & 0.653 & 0.646 & 0.702 & 0.739 & 0.690 & 0.732 \\
Constrained JSON (length norm, decision-only) & 0.161 & 0.497 & 0.640 & 0.706 & 0.795 & 0.654 & 0.689 & 0.725 & 0.738 & 0.691 & 0.734 \\
Constrained JSON (Google length norm) & 0.095 & 0.480 & 0.640 & 0.654 & 0.795 & 0.654 & 0.690 & 0.729 & 0.739 & 0.689 & 0.731 \\
Constrained JSON (Google length norm, decision-only) & 0.161 & 0.509 & 0.640 & 0.707 & 0.795 & 0.655 & 0.688 & 0.725 & 0.738 & 0.689 & 0.732 \\
Constrained JSON (probability tree) & 0.146 & \textbf{0.512} & 0.640 & 0.706 & 0.795 & 0.655 & 0.687 & 0.725 & 0.738 & 0.688 & 0.731 \\
Constrained JSON (ABC) & 0.056 & 0.391 & 0.602 & 0.694 & 0.790 & 0.592 & 0.650 & 0.710 & 0.714 & 0.659 & 0.704 \\
\bottomrule
\end{tabular}}
\end{table*}

\begin{table*}[t]
\centering
\caption{AUROC (mean score) across datasets, by model. \textbf{Bold} score represents the best performing for that model. Significance not tested as there are too few cells for the test to reach significance.}
\label{tab:best_model_auroc}
\resizebox{\textwidth}{!}{%
\begin{tabular}{lrrrrrrrrrrr}
\toprule
Method & Llama 3.2 1B & Llama 3.2 3B & Gemma 4 E2B & Gemma 4 E4B & Gemma 4 31B & Qwen2.5 7B & Qwen3 8B & Qwen3 14B & Qwen3 32B & Qwen3.5 4B & Qwen3.5 9B \\
\midrule
Unconstrained (no conf) & 0.500 & 0.500 & 0.500 & 0.500 & 0.500 & 0.500 & 0.500 & 0.500 & 0.500 & 0.500 & 0.500 \\
Unconstrained (no conf ABC) & 0.500 & 0.500 & 0.500 & 0.500 & 0.500 & 0.500 & 0.500 & 0.500 & 0.500 & 0.500 & 0.500 \\
Unconstrained (10 samples) & 0.581 & 0.654 & 0.525 & 0.559 & 0.515 & 0.539 & 0.532 & 0.535 & 0.581 & 0.673 & 0.659 \\
Unconstrained + P(True) & 0.473 & \textbf{0.754} & 0.685 & 0.719 & 0.767 & 0.724 & 0.726 & 0.751 & \textbf{0.786} & 0.749 & 0.759 \\
\midrule
Unconstrained JSON (no conf) & 0.500 & 0.500 & 0.500 & 0.500 & 0.500 & 0.500 & 0.500 & 0.500 & 0.500 & 0.500 & 0.500 \\
Unconstrained JSON (no conf ABC) & 0.500 & 0.500 & 0.500 & 0.500 & 0.500 & 0.500 & 0.500 & 0.500 & 0.500 & 0.500 & 0.500 \\
Unconstrained JSON verbal (pre) & 0.484 & 0.555 & 0.630 & 0.706 & 0.684 & 0.545 & 0.586 & 0.644 & 0.670 & 0.511 & 0.590 \\
Unconstrained JSON verbal (post) & 0.524 & 0.568 & 0.641 & 0.748 & 0.688 & 0.681 & 0.636 & 0.658 & 0.703 & 0.531 & 0.636 \\
Unconstrained JSON verbal (pre, Likert) & 0.531 & 0.550 & 0.581 & 0.657 & 0.642 & 0.594 & 0.555 & 0.641 & 0.618 & 0.576 & 0.587 \\
Unconstrained JSON verbal (post, Likert) & 0.522 & 0.579 & 0.603 & 0.686 & 0.673 & 0.620 & 0.582 & 0.653 & 0.682 & 0.559 & 0.668 \\
Unconstrained JSON verbal (pre, ABC) & 0.379 & 0.552 & 0.608 & 0.697 & 0.668 & 0.538 & 0.595 & 0.612 & 0.596 & 0.478 & 0.541 \\
Unconstrained JSON verbal (post, ABC) & 0.374 & 0.523 & 0.625 & 0.723 & 0.684 & 0.633 & 0.616 & 0.612 & 0.659 & 0.507 & 0.554 \\
\midrule
Constrained (greedy) & 0.500 & 0.500 & 0.500 & 0.500 & 0.500 & 0.500 & 0.500 & 0.500 & 0.500 & 0.500 & 0.500 \\
Constrained (total likelihood) & 0.606 & 0.734 & 0.719 & 0.787 & 0.763 & 0.736 & 0.739 & 0.774 & 0.772 & 0.786 & 0.798 \\
Constrained (total likelihood, decision-only) & 0.605 & 0.733 & 0.720 & 0.787 & 0.764 & 0.735 & 0.738 & 0.774 & 0.771 & 0.786 & 0.797 \\
Constrained (length norm) & \textbf{0.693} & 0.748 & 0.739 & 0.776 & 0.706 & 0.738 & 0.744 & \textbf{0.787} & 0.772 & 0.781 & 0.792 \\
Constrained (length norm, decision-only) & 0.683 & 0.743 & 0.715 & 0.779 & 0.736 & 0.731 & 0.723 & 0.765 & 0.762 & 0.785 & 0.793 \\
Constrained (Google length norm) & 0.637 & 0.739 & 0.727 & 0.787 & 0.745 & 0.738 & 0.739 & 0.777 & 0.773 & \textbf{0.789} & \textbf{0.802} \\
Constrained (Google length norm, decision-only) & 0.607 & 0.734 & 0.719 & 0.786 & 0.763 & 0.735 & 0.736 & 0.775 & 0.771 & 0.786 & 0.798 \\
Constrained (probability tree) & 0.605 & 0.731 & 0.720 & 0.786 & 0.764 & 0.734 & 0.737 & 0.774 & 0.772 & 0.786 & 0.797 \\
Constrained (ABC) & 0.656 & 0.735 & 0.698 & \textbf{0.791} & 0.762 & \textbf{0.756} & 0.745 & 0.758 & 0.766 & 0.764 & 0.779 \\
\midrule
Constrained JSON (greedy) & 0.500 & 0.500 & 0.500 & 0.500 & 0.500 & 0.500 & 0.500 & 0.500 & 0.500 & 0.500 & 0.500 \\
Constrained JSON (total likelihood) & 0.461 & 0.705 & \textbf{0.744} & 0.765 & \textbf{0.776} & 0.728 & 0.756 & 0.730 & 0.780 & 0.776 & 0.788 \\
Constrained JSON (total likelihood, decision-only) & 0.545 & 0.721 & \textbf{0.744} & 0.789 & \textbf{0.776} & 0.727 & 0.755 & 0.732 & 0.781 & 0.777 & 0.787 \\
Constrained JSON (length norm) & 0.496 & 0.743 & 0.736 & 0.762 & 0.724 & 0.735 & \textbf{0.770} & 0.696 & 0.709 & 0.745 & 0.766 \\
Constrained JSON (length norm, decision-only) & 0.612 & 0.729 & 0.733 & 0.778 & 0.746 & 0.721 & 0.736 & 0.706 & 0.753 & 0.761 & 0.779 \\
Constrained JSON (Google length norm) & 0.487 & 0.716 & \textbf{0.744} & 0.760 & 0.766 & 0.731 & 0.757 & 0.740 & 0.779 & 0.777 & 0.788 \\
Constrained JSON (Google length norm, decision-only) & 0.555 & 0.722 & 0.743 & 0.789 & 0.775 & 0.727 & 0.753 & 0.731 & 0.779 & 0.776 & 0.787 \\
Constrained JSON (probability tree) & 0.537 & 0.720 & \textbf{0.744} & 0.784 & \textbf{0.776} & 0.726 & 0.755 & 0.727 & 0.781 & 0.776 & 0.786 \\
Constrained JSON (ABC) & 0.620 & 0.729 & 0.733 & 0.782 & 0.772 & 0.744 & 0.762 & 0.776 & \textbf{0.786} & 0.764 & 0.776 \\
\bottomrule
\end{tabular}}
\end{table*}

\begin{table*}[t]
\centering
\caption{ECE (mean score) across datasets, by model. \textbf{Bold} score represents the best performing for that model. Significance not tested as there are too few cells for the test to reach significance.}
\label{tab:best_model_ece}
\resizebox{\textwidth}{!}{%
\begin{tabular}{lrrrrrrrrrrr}
\toprule
Method & Llama 3.2 1B & Llama 3.2 3B & Gemma 4 E2B & Gemma 4 E4B & Gemma 4 31B & Qwen2.5 7B & Qwen3 8B & Qwen3 14B & Qwen3 32B & Qwen3.5 4B & Qwen3.5 9B \\
\midrule
Unconstrained (no conf) & 0.755 & 0.513 & 0.355 & 0.272 & 0.199 & 0.319 & 0.294 & 0.269 & 0.251 & 0.280 & 0.249 \\
Unconstrained (no conf ABC) & 0.838 & 0.659 & 0.448 & 0.323 & 0.205 & 0.393 & 0.381 & 0.298 & 0.284 & 0.342 & 0.290 \\
Unconstrained (10 samples) & 0.495 & 0.386 & 0.344 & 0.253 & 0.196 & 0.302 & 0.281 & 0.257 & 0.231 & 0.221 & 0.201 \\
Unconstrained + P(True) & 0.331 & 0.323 & 0.333 & 0.225 & 0.187 & 0.273 & 0.266 & 0.229 & 0.190 & 0.179 & 0.158 \\
\midrule
Unconstrained JSON (no conf) & 0.727 & 0.485 & 0.344 & 0.275 & 0.200 & 0.331 & 0.291 & 0.267 & 0.259 & 0.282 & 0.248 \\
Unconstrained JSON (no conf ABC) & 0.840 & 0.579 & 0.396 & 0.306 & 0.202 & 0.396 & 0.342 & 0.285 & 0.280 & 0.331 & 0.287 \\
Unconstrained JSON verbal (pre) & 0.538 & 0.394 & 0.249 & 0.216 & 0.152 & 0.254 & 0.215 & 0.236 & 0.188 & 0.238 & 0.209 \\
Unconstrained JSON verbal (post) & 0.470 & 0.434 & 0.301 & 0.210 & 0.146 & 0.207 & 0.214 & 0.230 & 0.174 & 0.227 & 0.203 \\
Unconstrained JSON verbal (pre, Likert) & 0.494 & 0.270 & \textbf{0.180} & 0.194 & 0.120 & 0.162 & 0.174 & 0.188 & 0.146 & 0.180 & 0.160 \\
Unconstrained JSON verbal (post, Likert) & 0.386 & 0.281 & 0.191 & 0.188 & \textbf{0.107} & \textbf{0.141} & 0.138 & 0.192 & 0.121 & 0.152 & 0.120 \\
Unconstrained JSON verbal (pre, ABC) & 0.682 & 0.479 & 0.277 & 0.237 & 0.156 & 0.316 & 0.295 & 0.246 & 0.266 & 0.320 & 0.247 \\
Unconstrained JSON verbal (post, ABC) & 0.764 & 0.485 & 0.314 & 0.241 & 0.151 & 0.274 & 0.286 & 0.239 & 0.202 & 0.283 & 0.245 \\
\midrule
Constrained (greedy) & 0.780 & 0.510 & 0.361 & 0.272 & 0.199 & 0.319 & 0.294 & 0.269 & 0.251 & 0.279 & 0.249 \\
Constrained (total likelihood) & 0.356 & 0.238 & 0.329 & 0.228 & 0.192 & 0.265 & 0.267 & 0.242 & 0.204 & 0.143 & 0.138 \\
Constrained (total likelihood, decision-only) & 0.346 & 0.237 & 0.329 & 0.228 & 0.192 & 0.265 & 0.267 & 0.242 & 0.204 & 0.143 & 0.138 \\
Constrained (length norm) & \textbf{0.088} & 0.196 & 0.217 & 0.231 & 0.196 & 0.160 & 0.150 & \textbf{0.118} & 0.167 & 0.346 & 0.328 \\
Constrained (length norm, decision-only) & 0.140 & \textbf{0.094} & 0.312 & 0.198 & 0.187 & 0.217 & 0.243 & 0.223 & 0.164 & \textbf{0.066} & \textbf{0.053} \\
Constrained (Google length norm) & 0.268 & 0.175 & 0.320 & 0.214 & 0.190 & 0.250 & 0.260 & 0.236 & 0.191 & 0.103 & 0.107 \\
Constrained (Google length norm, decision-only) & 0.314 & 0.220 & 0.328 & 0.226 & 0.191 & 0.262 & 0.265 & 0.241 & 0.201 & 0.132 & 0.131 \\
Constrained (probability tree) & 0.335 & 0.236 & 0.329 & 0.228 & 0.192 & 0.266 & 0.267 & 0.242 & 0.204 & 0.142 & 0.138 \\
Constrained (ABC) & 0.357 & 0.189 & 0.414 & 0.270 & 0.199 & 0.316 & 0.345 & 0.270 & 0.217 & 0.125 & 0.148 \\
\midrule
Constrained JSON (greedy) & 0.772 & 0.486 & 0.364 & 0.285 & 0.198 & 0.335 & 0.297 & 0.279 & 0.258 & 0.289 & 0.256 \\
Constrained JSON (total likelihood) & 0.352 & 0.216 & 0.340 & 0.232 & 0.190 & 0.278 & 0.260 & 0.233 & 0.170 & 0.151 & 0.142 \\
Constrained JSON (total likelihood, decision-only) & 0.335 & 0.208 & 0.340 & 0.236 & 0.190 & 0.278 & 0.260 & 0.232 & 0.170 & 0.150 & 0.142 \\
Constrained JSON (length norm) & 0.198 & 0.228 & 0.220 & \textbf{0.157} & 0.197 & 0.191 & \textbf{0.113} & 0.184 & 0.379 & 0.387 & 0.411 \\
Constrained JSON (length norm, decision-only) & 0.188 & 0.098 & 0.322 & 0.198 & 0.184 & 0.221 & 0.225 & 0.195 & \textbf{0.091} & 0.123 & 0.110 \\
Constrained JSON (Google length norm) & 0.273 & 0.145 & 0.332 & 0.216 & 0.187 & 0.262 & 0.248 & 0.225 & 0.138 & 0.112 & 0.106 \\
Constrained JSON (Google length norm, decision-only) & 0.304 & 0.193 & 0.338 & 0.233 & 0.190 & 0.274 & 0.257 & 0.230 & 0.163 & 0.140 & 0.134 \\
Constrained JSON (probability tree) & 0.332 & 0.206 & 0.340 & 0.237 & 0.190 & 0.278 & 0.260 & 0.234 & 0.170 & 0.149 & 0.141 \\
Constrained JSON (ABC) & 0.198 & 0.190 & 0.374 & 0.261 & 0.196 & 0.312 & 0.309 & 0.256 & 0.207 & 0.155 & 0.155 \\
\bottomrule
\end{tabular}}
\end{table*}

\begin{table*}[t]
\centering
\caption{Top-label Brier (mean score) across datasets, by model. \textbf{Bold} score represents the best performing for that model. Significance not tested as there are too few cells for the test to reach significance.}
\label{tab:best_model_binary_brier}
\resizebox{\textwidth}{!}{%
\begin{tabular}{lrrrrrrrrrrr}
\toprule
Method & Llama 3.2 1B & Llama 3.2 3B & Gemma 4 E2B & Gemma 4 E4B & Gemma 4 31B & Qwen2.5 7B & Qwen3 8B & Qwen3 14B & Qwen3 32B & Qwen3.5 4B & Qwen3.5 9B \\
\midrule
Unconstrained (no conf) & 0.755 & 0.513 & 0.355 & 0.272 & 0.199 & 0.319 & 0.294 & 0.269 & 0.251 & 0.280 & 0.249 \\
Unconstrained (no conf ABC) & 0.838 & 0.659 & 0.448 & 0.323 & 0.205 & 0.393 & 0.381 & 0.298 & 0.284 & 0.342 & 0.290 \\
Unconstrained (10 samples) & 0.456 & 0.372 & 0.344 & 0.254 & 0.196 & 0.302 & 0.282 & 0.258 & 0.231 & 0.226 & 0.206 \\
Unconstrained + P(True) & 0.308 & 0.305 & 0.335 & 0.232 & 0.187 & 0.277 & 0.267 & 0.231 & 0.198 & 0.206 & 0.182 \\
\midrule
Unconstrained JSON (no conf) & 0.727 & 0.485 & 0.344 & 0.275 & 0.200 & 0.331 & 0.291 & 0.267 & 0.259 & 0.282 & 0.248 \\
Unconstrained JSON (no conf ABC) & 0.840 & 0.579 & 0.396 & 0.306 & 0.202 & 0.396 & 0.342 & 0.285 & 0.280 & 0.331 & 0.287 \\
Unconstrained JSON verbal (pre) & 0.505 & 0.398 & 0.270 & 0.229 & 0.162 & 0.306 & 0.248 & 0.268 & 0.214 & 0.256 & 0.221 \\
Unconstrained JSON verbal (post) & 0.439 & 0.441 & 0.307 & 0.222 & 0.158 & 0.238 & 0.233 & 0.263 & 0.196 & 0.244 & 0.213 \\
Unconstrained JSON verbal (pre, Likert) & 0.489 & 0.313 & 0.253 & 0.235 & 0.145 & 0.250 & 0.237 & 0.227 & 0.215 & 0.243 & 0.190 \\
Unconstrained JSON verbal (post, Likert) & 0.355 & 0.323 & 0.251 & 0.218 & \textbf{0.140} & \textbf{0.214} & 0.204 & 0.214 & 0.178 & 0.228 & 0.173 \\
Unconstrained JSON verbal (pre, ABC) & 0.594 & 0.469 & 0.293 & 0.246 & 0.167 & 0.352 & 0.305 & 0.277 & 0.287 & 0.331 & 0.258 \\
Unconstrained JSON verbal (post, ABC) & 0.738 & 0.491 & 0.316 & 0.252 & 0.162 & 0.298 & 0.297 & 0.268 & 0.225 & 0.295 & 0.256 \\
\midrule
Constrained (greedy) & 0.780 & 0.510 & 0.361 & 0.272 & 0.199 & 0.319 & 0.294 & 0.269 & 0.251 & 0.279 & 0.249 \\
Constrained (total likelihood) & 0.297 & 0.258 & 0.329 & 0.231 & 0.192 & 0.270 & 0.268 & 0.243 & 0.208 & 0.179 & 0.167 \\
Constrained (total likelihood, decision-only) & 0.288 & 0.258 & 0.329 & 0.231 & 0.192 & 0.271 & 0.268 & 0.243 & 0.208 & 0.178 & 0.167 \\
Constrained (length norm) & \textbf{0.128} & 0.253 & \textbf{0.250} & 0.233 & 0.185 & 0.215 & 0.194 & \textbf{0.177} & 0.177 & 0.322 & 0.294 \\
Constrained (length norm, decision-only) & 0.151 & 0.199 & 0.311 & 0.208 & 0.188 & 0.239 & 0.247 & 0.226 & 0.183 & \textbf{0.155} & \textbf{0.142} \\
Constrained (Google length norm) & 0.216 & 0.227 & 0.320 & 0.219 & 0.190 & 0.259 & 0.261 & 0.237 & 0.198 & 0.164 & 0.153 \\
Constrained (Google length norm, decision-only) & 0.257 & 0.248 & 0.327 & 0.229 & 0.192 & 0.268 & 0.266 & 0.242 & 0.205 & 0.174 & 0.163 \\
Constrained (probability tree) & 0.278 & 0.258 & 0.329 & 0.231 & 0.192 & 0.271 & 0.268 & 0.243 & 0.208 & 0.178 & 0.167 \\
Constrained (ABC) & 0.274 & \textbf{0.186} & 0.410 & 0.270 & 0.199 & 0.309 & 0.339 & 0.269 & 0.226 & 0.189 & 0.183 \\
\midrule
Constrained JSON (greedy) & 0.772 & 0.486 & 0.364 & 0.285 & 0.198 & 0.335 & 0.297 & 0.279 & 0.258 & 0.289 & 0.256 \\
Constrained JSON (total likelihood) & 0.237 & 0.262 & 0.338 & 0.243 & 0.191 & 0.283 & 0.261 & 0.240 & 0.189 & 0.190 & 0.177 \\
Constrained JSON (total likelihood, decision-only) & 0.270 & 0.252 & 0.338 & 0.239 & 0.191 & 0.283 & 0.261 & 0.238 & 0.188 & 0.189 & 0.177 \\
Constrained JSON (length norm) & 0.144 & 0.269 & 0.253 & \textbf{0.204} & 0.186 & 0.231 & \textbf{0.185} & 0.218 & 0.352 & 0.372 & 0.370 \\
Constrained JSON (length norm, decision-only) & 0.165 & 0.208 & 0.320 & 0.212 & 0.185 & 0.245 & 0.234 & 0.214 & \textbf{0.165} & 0.178 & 0.164 \\
Constrained JSON (Google length norm) & 0.176 & 0.232 & 0.330 & 0.240 & 0.188 & 0.270 & 0.250 & 0.231 & 0.174 & 0.174 & 0.164 \\
Constrained JSON (Google length norm, decision-only) & 0.250 & 0.244 & 0.336 & 0.237 & 0.190 & 0.279 & 0.258 & 0.236 & 0.185 & 0.185 & 0.174 \\
Constrained JSON (probability tree) & 0.264 & 0.252 & 0.338 & 0.240 & 0.191 & 0.282 & 0.261 & 0.240 & 0.188 & 0.189 & 0.177 \\
Constrained JSON (ABC) & \textbf{0.128} & 0.237 & 0.366 & 0.261 & 0.196 & 0.305 & 0.306 & 0.254 & 0.216 & 0.196 & 0.186 \\
\bottomrule
\end{tabular}}
\end{table*}

\clearpage
\section{Prompts}
\label{sec:prompts}
\begin{promptbox}{Fully rendered prompt (FEVER, plaintext label, Gemma~4 31B chat template)}
<bos><|turn>system
# CONTEXT
You are an expert fact-checker. You specialize in analyzing claims with provided evidence and deciding whether the evidence supports the claim, refutes the claim, or there is not enough evidence to make a clear decision.

# STYLE & TONE
Favor complete brevity. You need not respond with any reasoning, just the answer.

# FORMAT
* You will be provided with the claim inside <claim></claim> tags.
* Evidence will be provided within the <evidence></evidence> tags.
* You will be provided with each set of evidence inside <evidence_set></evidence_set> tags.
* Within the evidence tags, you will be provided each piece of evidence inside <evidence_piece title=[TITLE]></evidence_piece> tags, where the title of the wiki pages article is provided in the first tag.
* Respond with ONLY the answer from one of these options:
    - SUPPORTS
    - REFUTES
    - NOT ENOUGH INFO<turn|>
<|turn>user
# OBJECTIVE
Classify whether the provided evidence SUPPORTS or REFUTES the given claim, or whether there is NOT ENOUGH INFO. Rely ONLY on the evidence provided and use any combination of the pieces of evidence provided.

# INPUT
<claim>
  Aruba is the only ABC Island.
</claim>

<evidence>
  <evidence_set>
    <evidence_piece title=ABC islands (Lesser Antilles)>
      The ABC islands are the three western-most islands of the Leeward Antilles in the Caribbean Sea that lie north of Falcón State, Venezuela.
    </evidence_piece>
  </evidence_set>
  <evidence_set>
    <evidence_piece title=ABC islands (Lesser Antilles)>
      From west to east they are, in order Aruba, Curaçao, and Bonaire.
    </evidence_piece>
  </evidence_set>
</evidence>

# ANSWER
Answer (choose between SUPPORTS, REFUTES, NOT ENOUGH INFO):<turn|>
<|turn>model
<|channel>thought
<channel|>
\end{promptbox}

\begin{promptbox}{Fully rendered prompt (Banking77, JSON verbal (post), Gemma~4 31B chat template)}
<bos><|turn>system
# CONTEXT
You are an expert customer support analyst for online banking. You specialize in analyzing customer queries and classifying them into one of 77 intent classes.

# STYLE & TONE
Favor complete brevity. You need not respond with reasoning, just the answer.

# FORMAT
* The customer query will be provided inside <query></query> tags.
* Your response MUST be in JSON format, with the intent label FIRST and confidence SECOND.
* The "label" key in your answer must be only ONE of the following intents:
  - activate_my_card
  - age_limit
  - ...
* The "confidence" key in your answer MUST be provided as a percentage to reflect your confidence in the label you are about to provide.
* Return ONLY the JSON object. NO OTHER TEXT.
* Example format:
  {
    "label": "[INTENT]",
    "confidence": "X%"
  }<turn|>
<|turn>user
# OBJECTIVE
Provide confidence about your ability to classify the intent of the customer query and classify the customer's intent with their query. Choose the most specific intent that best matches their query.

# INPUT
<query>
  How do I locate my card?
</query>

# ANSWER
JSON output with the intent label FIRST and confidence SECOND ({"label": "[choose between activate_my_card, age_limit, ...]", "confidence": "[PERCENTAGE]%"}):<turn|>
<|turn>model
<|channel>thought
<channel|>
\end{promptbox}

\begin{promptbox}{Fully rendered prompt (CLINC150, JSON verbal (pre, Likert), Gemma~4 31B chat template)}
<bos><|turn>system
# CONTEXT
You are an expert virtual assistant analyst. You specialize in analyzing user queries and classifying them into one of 150 intent classes.

# STYLE & TONE
Favor complete brevity. You need not respond with reasoning, just the answer.

# FORMAT
* The user query will be provided inside <query></query> tags.
* Your response MUST be in JSON format, with confidence FIRST and the intent label SECOND.
* The "confidence" key in your answer MUST be one of the following labels to reflect your confidence in the label you are about to provide:
  - Not at all confident
  - Slightly confident
  - Moderately confident
  - Very confident
  - Completely confident
* The "label" key in your answer must be only ONE of the following intents:
  - accept_reservations
  - account_blocked
  - alarm
  - ...
* Example format:
  {
    "confidence": "Not at all confident" | "Slightly confident" | "Moderately confident" | "Very confident" | "Completely confident",
    "label": "[INTENT]"
  }<turn|>
<|turn>user
# OBJECTIVE
Classify the user's intent with their query. Choose the most specific intent that best matches their query. If the query does not belong to any listed intent, answer oos.

# INPUT
<query>
  how would you say fly in italian
</query>

# ANSWER
JSON output with confidence FIRST and intent label SECOND ({"confidence": "[choose between Not at all confident, Slightly confident, Moderately confident, Very confident, Completely confident]", "label": "[choose between restaurant_reviews, nutrition_info, ...]"}):<turn|>
<|turn>model
<|channel>thought
<channel|>
\end{promptbox}

\begin{promptbox}{Fully rendered prompt (SST-5, plaintext label (P(True)), Gemma~4 31B chat template)}
<bos><|turn>system
# CONTEXT
You are an expert sentiment analysis analyst for movie reviews. You specialize in analyzing review excerpts and classifying them into one of 5 sentiment classes.

# STYLE & TONE
Favor complete brevity. You need not respond with any reasoning, just the answer.

# FORMAT
* You will be provided with the instructions given to the annotator and the annotator's response.
* You must respond with whether the annotator is correct or not ("True" or "False").
* You must ONLY respond with True or False, no other text.<turn|>
<|turn>user
# OBJECTIVE
Judge whether the response provided by an annotator is correct (True or False). They were tasked to classify the sentiment of the review excerpt, choosing the sentiment that best matches the overall tone from the list of 5 valid sentiments.

# INPUT & ANNOTATOR RESPONSE
"""
# INPUT
# CONTEXT
You are an expert sentiment analysis analyst for movie reviews. You specialize in analyzing review excerpts and classifying them into one of 5 sentiment classes.

# STYLE & TONE
Favor complete brevity. You need not respond with reasoning, just the answer.

# FORMAT
* The review excerpt will be provided inside <review></review> tags.
* Respond with ONLY the answer from the list of valid sentiments provided:
  - very negative
  - negative
  - neutral
  - positive
  - very positive


# OBJECTIVE
Classify the sentiment of the review excerpt. Choose the sentiment that best matches the overall tone.

# INPUT
<review>
  no movement , no yuks , not much of anything .
</review>

# ANSWER
Answer (choose from: very negative, negative, neutral, positive, very positive):


# RESPONSE
negative
"""

# ANSWER
Correct? (True | False):<turn|>
<|turn>model
<|channel>thought
<channel|>
\end{promptbox}

\begin{promptbox}{Fully rendered prompt (20NG, JSON output (ABC, with prefill for logit method), Gemma~4 31B chat template)}
<bos><|turn>system
# CONTEXT
You are an expert text classification analyst for online discussion forums. You specialize in analyzing newsgroup posts and classifying them into one of 20 topic categories.

# STYLE & TONE
Favor complete brevity. You need not respond with reasoning, just the answer.

# FORMAT
* The newsgroup post will be provided inside <post></post> tags.
* Your response MUST be in JSON format, containing only the category label.
* Respond with ONLY the answer from the list of valid categories provided:
  - A (alt.atheism)
  - B (comp.graphics)
  - C (comp.os.ms-windows.misc)
  - D (comp.sys.ibm.pc.hardware)
  - E (comp.sys.mac.hardware)
  - F (comp.windows.x)
  - G (misc.forsale)
  - H (rec.autos)
  - I (rec.motorcycles)
  - J (rec.sport.baseball)
  - K (rec.sport.hockey)
  - L (sci.crypt)
  - M (sci.electronics)
  - N (sci.med)
  - O (sci.space)
  - P (soc.religion.christian)
  - Q (talk.politics.guns)
  - R (talk.politics.mideast)
  - S (talk.politics.misc)
  - T (talk.religion.misc)
* Example format:
  {
    "label": "[CATEGORY_LETTER]"
  }<turn|>
<|turn>user
# OBJECTIVE
Classify the newsgroup post into its topic category. Choose the LETTER that represents the most specific category that best matches the post's content.

# INPUT
<post>
  I am a little confused on all of the models of the 88-89 bonnevilles.
I have heard of the LE SE LSE SSE SSEI. Could someone tell me the
differences are far as features or performance. I am also curious to
know what the book value is for prefereably the 89 model. And how much
less than book value can you usually get them for. In other words how
much are they in demand this time of year. I have heard that the mid-spring
early summer is the best time to buy.
</post>

# ANSWER
JSON output containing only the category label ({"label": "choose between A, B, C, D, E, F, G, H, I, J, K, L, M, N, O, P, Q, R, S, T"}):<turn|>
<|turn>model
<|channel>thought
<channel|>```json
{"label": "
\end{promptbox}

\end{document}